\documentclass[11pt,a4paper]{article}
\usepackage{linguex}
\usepackage{graphicx}
\usepackage[hyperref]{ranlp2023}
\usepackage{times}
\usepackage{latexsym}

\usepackage{microtype}

\aclfinalcopy 

\title{Using ChatGPT as a CAT tool in Easy Language translation}

\author{Silvana Deilen \qquad Sergio Hernández Garrido,\\ \bf{Ekaterina Lapshinova-Koltunski} \qquad\bf{Christiane Maaß}\\\\
  University of Hildesheim \\
     \texttt{\{deilen,hernandezs,lapshinovakoltun,maassc\}@uni-hildesheim.de}}

\date{}

\begin{document}
\maketitle
\begin{abstract}
 This study sets out to investigate the feasibility of using ChatGPT to translate citizen-oriented administrative texts into German Easy Language, a simplified, controlled language variety that is adapted to the needs of people with reading impairments. We use ChatGPT to translate selected texts from websites of German public authorities using two strategies, i.e. linguistic and holistic. We analyse the quality of the generated texts based on different criteria, such as correctness, readability, and syntactic complexity. The results indicated that the generated texts are easier than the standard texts, but that they still do not fully meet the established Easy Language standards. Additionally, the content is not always rendered correctly.

\end{abstract}

\section{Introduction}\label{sec:intro}


Generative Pre-trained Transformer (GPT) models show remarkable advances not only in natural language generation \cite{BrownEtAl2020}, but also in automated translation~\cite{hendy2023}. However, their performance in specific machine translation tasks has not been yet extensively explored. While the online ChatGPT is also known for its ability to translate texts from one language to another (so-called interlingual translation), so far very little is known about its ability to translate texts from standard language into a complexity-reduced language variety of the same language (so-called intralingual translation). In this study, we investigate the feasibility of using ChatGPT to translate citizen-oriented administrative texts from public agencies into Easy Language for people with reading impairments. The aim of our study is twofold: first, to answer the question of whether and to what extent large language models like ChatGPT are able to generate translations from standard German into Easy Language and second, to determine whether a holistic or a text-based approach leads to a more comprehensible output. 

In our study, we aim to test whether ChatGPT is fit to be used as a tool for translating texts into German Easy Language. We selected a number of texts and ordered the tool to perform several translation tasks with two different strategies: linguistic level dependent and holistic. In this paper, we present the results of the qualitative and quantitative data from the analysis and on their basis, we draw the first conclusions on the usability of ChatGPT as an Easy Language CAT\footnote{Computer-aided translation} tool.

The remainder of this paper is structured as follows. In Section~\ref{sec:easylang}, we outline the relevance of the demand for the current situation with Easy Language in Germany. In Section~\ref{sec:related}, we summarize the existing related work. In Section~\ref{sec:design}, we describe the data used in this study and outline our research design. In Section~\ref{sec:results}, we present our main results and lastly, in Section~\ref{sec:summarydiscussion}, we discuss our main findings and suggest promising directions for future research.

\section{Easy Language in Germany}\label{sec:easylang}

In Germany, the Federal Act on Equal Opportunities of Persons with Disabilities (Behindertengleichstellungsgesetz [BGG] 2016/2018) states that public authorities must on request explain official notices, general rulings, public-law contracts and forms in Easy Language to individuals with intellectual or psychological disabilities. However, since many German public authorities still do not provide sufficient information in Easy Language, these rights to information in Easy Language are often disregarded.
One of the main reasons for the lack of information in Easy Language is that translating texts into Easy Language is very time-consuming and costly and requires professional training (cf. \citealt{maass2020easy}, \citealt{hansen2020technologies}). 


Consequently, the problem is twofold: On the one hand there is an increasing demand for texts in Easy Language and, due to the legal situation, a significantly rising text volume to be translated \cite{rink2019rechtskommunikation}, and on the other hand, people who are responsible for translating texts into Easy Language often lack sufficient translation experience and expertise  \cite{maass2020easy}. At the moment, Easy Language translations are not always carried out by academically trained translators but also often by persons without academic translation training, such as employees of public authorities or organizations or social education workers, who have access to the Easy Language target groups but who do not have “the necessary text expertise and adequate formal training in intralingual translation“ (\citealt[p. 121] {hansen2020technologies}). This lack of professional academic training leads to a heterogeneous and often very poor text quality. This is problematic for several reasons:  Firstly, texts in Easy Language are only functional when they are of high quality. This is especially true for target groups with special communication needs who may not be able to understand poorly written texts (for an overview of the Easy Language target groups see \citealt{bredel2016leichte}). In addition, poorly written texts can also negatively affect the public image of Easy Language and may even stigmatize its users (cf. \citealt{maass2020easy}). 
Furthermore, texts of poor quality are detrimental to the development of machine translation systems for Easy Language, because the successful training of such systems requires a large corpus of rule-consistent high quality Easy Language translations (cf. \citealt{hansen2020technologies}). Thus, texts of poor quality hinder the compilation of such a corpus and therefore slow down the advancement of automatic text simplification systems that can be used for intralingual translation. 

However, due to the significantly rising text volume and the lack of professional translators, the need for technological assistance is obvious. As texts in Easy Language are based on defined rules on the word, sentence and text level, Easy Language is often treated as a controlled language (cf. \citealt{hansen2020technologies}). This in turn means that, seen from a theoretical perspective, they offer a high automation potential. Yet, due to the above-mentioned reasons,  texts in Easy Language are mostly translated manually.

As large language models (LLMs) like ChatGPT are trained on large amounts of data from the internet and use this data to generate new content, it is conceivable that LLMs like ChatGPT are able to reduce the complexity of a text by applying strategies of lexical and syntactic simplification. Seen from a theoretical perspective, it therefore seems plausible that LLMs have the potential to convert a source text into a text version that is easier to read and understand. However, at this time we are not aware of any studies that evaluate the feasibility of such systems for intralingual translation.

\section{Related work}\label{sec:related}
\subsection{Easy German}\label{ssec:related:leichtesprache}
Easy German has become a subject of scientific research since 2014~\cite{maass2014leichte} with rapidly growing output of publications in the following years for German and other national Easy varieties. The studies point in two basic directions: studies on text qualities and possible  barriers in various forms of communication on the on the one side (see, for example, \citealt{rink2019rechtskommunikation} for legal communication in Easy Language) and studies on comprehensibility and recall by different target groups on the other (see, for example, \citealt{gutermuth2020leichte} and \citealt{deilen2021optische}). For an overview on the situation of Easy Languages in Europe see \citealt{lindholm2021handbook} and, specifically, the chapter on Easy German \citealt{maass2021easy}).

\subsection{Automatic text simplification for Easy and Plain Language}\label{ssec:related:textsimplification}
Even though there are many previous studies on automatic text simplification methods that aim to automatically convert a text into another text that is easier to understand but ideally conveys the same message as the source text (cf. \citealt{saggion2017applications}), the role of automation and CAT tools for Easy Language translation is still a major research desideratum. Easy and Plain language display different grades of comprehensibility and address differing target groups that need accessible communication to participate in various fields of society \cite{bredel2016leichte}. \citet{maass2014leichte} were the first to discuss the potentials of computer-aided translation tools for Easy Language translation. In their 2020 paper, \citet{hansen2020technologies} reconsidered and extended these potentials and published them for an international scientific community. Both papers show that intralingual terminology management comes with some challenges because, in contrast to interlingual translation, in Easy Language translation the description, explanation and definition of a concept has to be made explicit in the text and cannot be hidden in the termbase. Furthermore, when it comes to intralingual sentence alignment, there is usually no 1:1 correspondence between source text and target text. This is due to sentence compression or splitting strategies, additional explanations, or the shifting of the order of information in the source text. This in turn means that the alignment process has to be done or corrected manually by the translator, which increases the workload. With regard to the use of translation memories, they suggest lowering the threshold value for fuzzy matches, because in intralingual translation also matches below 70\% (which is the common threshold in interlingual translation) can be used as a template and can therefore be useful for the translator. As a consequence, they conclude that intralingual terminology management is feasible, but requires specific adaptations of the best practices. Likewise, \citet{welch2019easy} conclude that the structure of common interlingual terminology systems is too restrictive to be used in Easy Language translation. After listing the requirements for an intralingual termbase they therefore propose a theoretical set-up, additional fields and features that would be needed in an Easy Language terminology tool. However, to our knowledge such a tool still does not exist.

Although existing studies in automatic text simplification operating with deep learning methods (see e.g. \citealt{SheangSaggion2021,maddela-etal-2021-controllable,martin-etal-2020-controllable} amongst others) also aim at textual accessibility, most of them do not  consider the needs of target audience. \citet{scarton-specia-2018-learning} did present an approach for automatic text simplification that makes use of the Newsela corpus\footnote{\url{https://newsela.com/data}}. This corpus was built for various target audiences with each corpus article being labeled with a grade level and having also various simplified versions. The authors showed that using such target audience oriented data helped to build better models than general purpose ones. However, such models do not necessarily reflect the specificities of Easy Language. 

To our knowledge, \citet{sauberli2020benchmarking} were the first to adapt neural models to the features of German Easy Language. Their models were able to implement some specificities of Easy Language, such as choosing basic words or shortening sentences. However, despite these achievements, they also showed that in most cases the content was not preserved or contained wrong details. As their analysis also revealed that in most cases, the sentences were not significantly easier than the original sentences, they conclude that a larger parallel corpus is needed to successfully train an automatic text simplification system for German Easy Language. 

\citet{spring2021exploring} expanded the corpus used by \citet{sauberli2020benchmarking} and developed a sentence-based machine translation approach to automatically simplify standard German into different simplification levels of the Common European Framework of References for Languages (CEFR). To tackle the above-mentioned alignment problems, they used the Sentence Alignment Tools Evaluation Framework (SATEF), which allows for n:m alignments, meaning that one alignment segment can consist of a varying number of sentences in the source and target text. Alignment issues were also addressed by
\citet{KoppEtAl2023} who developed a translation memory for non-professional intralingual translators in the field of public administration. Its main functionality lies in the assistance in the creation of alignment corpora in standard language and Easy Language by using automatic alignment algorithms. This translation memory serves in the short term as a database with aligned text passages that support the translation process into Easy Language. In the long term, the created corpora can serve as high quality data to train AI for intralingual machine translation purposes. 

However, both~\citet{sauberli2020benchmarking}
and ~\citet{spring2021exploring} showed that 
existing models tend to copy the source segments. The latter were able to reduce the copying behavior of the text simplification models by applying different pretraining and fine-tuning strategies and by adding copy labels. As their simplification models mostly outperformed the baseline models in terms of the BLEU score~\cite{PapineniEtal2002bleu} and SARI~\cite{XuEtal2016optimizing}, their study showed that pretrained and fine-tuning NMT models is a promising approach to German automatic text simplification. \citet{anschutz2023language} also used fine-tuning for five pre-trained language models for German Easy Language. They found that both in terms of models’ perplexities and readability of the output the fine-tuned models showed better conformity to the linguistic features and structure of German Easy Language than the original versions of the models. Therefore, their study revealed that it is possible to train models to adapt to the style of German Easy Language. They conclude that even though the generated output might not be used by the target groups directly, it might serve as a draft for professional German Easy Language translators and might thus, similarly to post-editing in interlingual translation, reduce their workload.

Although the above mentioned studies show advances in applying neural models to Easy Language, none of them evaluated the outcome generated by an already existing, non-self-trained model.

\subsection{LLMs / ChatGPT for translation tasks}\label{ssec:related:chagpt}

As already mentioned in Section~\ref{sec:intro} above, GPT models have been successfully tested for automated translation in various tasks. For instance, \citet{hendy2023} analysed performance of three GPT models (including ChatGPT) for different translation directions showing that such models achieved competitive translation quality for high resource languages. 
\citet{kocmi2023large} used GPT models to test if these can be applied for automatic translation quality assessment. 
The authors showed that their quality assessment scheme correlates with larger models only. Interestingly, their method for translation quality assessment only works with GPT 3.5 and larger models. They also showed that 
the least constrained template achieved the best performance in this analysis.

Apart from overall translation tasks, ChatGPT has been tested for handling specific linguistic phenomena, e.g. translation of coreference chains, ellipsis, terminology and other lexical issues and especially ambiguous constructions~\cite{CastilhoEtAl2023}. ChatGPT turned to deal better with context-related issues than other MT engines under analysis and also suggest creative translation solutions. 

To our knowledge, none of the existing studies has addressed the performance of ChatGPT for intralingual translation tasks, specifically for German Easy Language. The only study known to us that addresses readability, which is one of the features we analyse, is~\citet{PuDemberg2023}. The authors compare reading difficulty of the ChatGPT outputs with human-written texts. Their results show that although ChatGPT-generated sentences for experts showed greater complexity than for layperson, the magnitude of the difference in the reading difficulty scores between the two types of texts (for experts vs. layperson) was much smaller than that observed in human-generated texts. 

\section{Research Design}\label{sec:design}

\subsection{Data collection} \label{sec:datacollection}
To test the chatbot ChatGPT\footnote{Our study was conducted in April 2023, i.e., the results are based on GPT-3.5, the latest free version of ChatGPT available at the time of writing.} for intralingual translation into German Easy Language, we used twenty texts from three different websites of German public authorities. Each text contained between 179 and 672 words. The texts contained information about different citizen-oriented topics, such as how to report lost and found items, how to take parental leave, or how to obtain a criminal record certificate.

In our study, we tested two different approaches: As human translators usually follow a holistic approach when translating a text, our first approach corresponds to a natural translation strategy. However, as German Easy Language is a controlled language that is characterized by specific rules on text, sentence, and word level, it is also conceivable that simplifying the linguistic levels separately improves the machine generated output. In our second approach, the so-called linguistic level dependent approach, we therefore adapted our prompts to the text, sentence and word level respectively. 

Starting with the holistic approach, we first asked the tool to translate the following text into German Easy Language. However, when looking at the generated output, it quickly became clear that the texts did not follow the common German Easy Language rules and were still too complex. For example, the independent clause-only principle was violated and the texts still contained complex nominal phrases. Therefore, in a second step we requested ChatGPT to make the text easier. This request was formulated twice.

Afterwards, we tested the second approach. In this approach, we tried to simplify the source texts step by step, according to the strategies that are applied in Easy Language translation. We differentiated between simplifying strategies on text level, sentence level and word level. Starting from the text level, we first asked ChatGPT to reformulate the text but to leave out unimportant information. In a second step, we requested the tool to reformulate the text without compound sentences and with simple syntactic structures. In a third step, we requested ChatGPT to add explanations of difficult words in the text. In our analysis, we only considered the final outputs of the two approaches, i.e., the version the tool generated after each of the respective last query. Table~\ref{tab:corpus} provides an overview of the resulting subcorpora under analysis\footnote{The analysed data is available under \url{https://github.com/katjakaterina/chatgpt4easylang}.}. They include source texts (S), texts generated with the holistic approach (H), and the texts generated with the linguistic approach (L).

\begin{table}[ht!]
    \centering
    \begin{tabular}{|l|r|}
    \hline
 \bf    subcorpus & \bf tok \\
 \hline
   source (S) & 8.919 \\
holistic (H) & 2.707 \\
linguistic (L) & 5.950 \\
\hline
\bf total & \bf 17.576 \\
\hline
    \end{tabular}
    \caption{Corpus statistics in tokens (tok)}
    \label{tab:corpus}
\end{table}


Then, we compared the three supcorpora using three different criteria: The first criterion was the correctness of the content (see~\ref{ssec:correctness}) applied to the H and L subcorpora only, the second criterion was the readability of the generated output (see~\ref{ssec:readability}), and the third criterion was the syntactic complexity of the texts (see~\ref{ssec:syntcomplexity}).

\subsection{Data analysis}
\label{sec:dataanalysis}

\subsubsection{Correctness}\label{ssec:correctness}
In our analysis, we first evaluated whether the content of the generated texts is correct. The evaluation was done according to the four-eyes principle, e.g., the correctness of each text was evaluated independently by two people. In case of discrepancies, the respective text was reviewed and discussed in plenary until a unanimous decision was reached.
As we know that like other LLMs, ChatGPT suffers from hallucination issues in the context of logical reasoning \cite{BangEtAl2023}, we expect to find some incorrect contents in the H and L subcorpora.

\subsubsection{Readability}\label{ssec:readability}
Secondly, we compared the comprehensibility of the different approaches. The comprehensibility was assessed by TextLab, a software that determines text-comprehensibility based on the Hohenheim Comprehensibility Index (HIX). The HIX is a meta index that calculates the readability of a text taking into account the four major readability formulas common in German Easy Language Research~\cite[p. 61ff]{bredel2016leichte}. They include the Amstad index, the simple measure of gobbledygook (G-SMOG) index, the Vienna non-fictional text formula (W-STX) and the readability index (LIX), with an index of 0 indicating an extremely low comprehensibility and an index of 20 an extremely high comprehensibility (for further details see: \url{https://klartext.uni-hohenheim.de/hix}). To evaluate whether a text can be classified as a German Easy Language text, we used a predefined benchmark for German Easy Language, according to which Easy Language texts should have a HIX of at least 18 points (cf. \citealt [p. 77] {rink2019rechtskommunikation}).

\subsubsection{Syntactic complexity}\label{ssec:syntcomplexity}
We operationalise syntactic complexity as a distribution of specific syntactic relations, i.e. specific clauses. We automatically identify syntactic relations using dependency parsing that we obtained with the Stanford NLP Python Library Stanza (v1.2.1)\footnote{\url{https://stanfordnlp.github.io/stanza/index.html}} with all the models pre-trained on the Universal Dependencies v2.5 datasets. Our list of selected structural categories include the following: acl (adnominal clause or clausal modifier of noun), advcl (adverbial clause modifier), ccomp (clausal component), csubj (clausal subject), xcomp (open clausal element) and parataxis (parataxis relation). They are all listed under the clause dependents\footnote{\url{https://universaldependencies.org/u/dep/}} in the Universal Dependency (see \citealt{de2021universal}for more details) definition. The occurrence of these categories is collected and analysed across the three subcorpora under analysis. We assume that the higher the number of these dependency relations in the corpus, the more complex the texts contained in these subcorpora are.

\section{Results}\label{sec:results}

\subsection{Correctness}
\label{ssec:res:correctness}
Analyzing the correctness of the content revealed that, altogether, 37.5\% of the generated texts were content-wise correct. In 62.5\% however, the text contained at least one incorrect piece of information. When looking at the two approaches separately, we found that from the holistic output, 80\% of the texts were marked as incorrect, whereas from the linguistic level dependent output,  45\% of the texts were classified as incorrect. An example of an incorrect translation is illustrated in (1).

\begin{quote}

1a. \textit{Bis zum 18. Lebensjahr ist auch der gesetzliche Vertreter antragsbefugt. [Up to the age of 18, the legal representative is also authorised to file the application]} (16S)

1b. \textit{Wenn man unter 18 ist, kann es nur von einem gesetzlichen Vertreter beantragt werden. [If you are under 18, it can only be filed by a legal representative]} (16H)
    
\end{quote}

While the source sentence in (1a) means that both a person under 18 and her/his legal representative are authorised to file the application, the translation output in (1b) means that only the representative can do so.

\subsection{Readability}
\label{ssec:res:readability}

\begin{figure}[h!]
   \centering 
    \vspace{-1mm}
\includegraphics[width=0.45\textwidth]{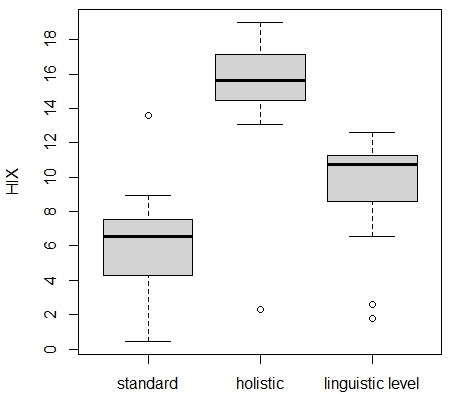}
    \vspace{-1.5mm}
    \caption{HIX values of the source text and the two simplified variants under analysis.}
     \label{fig:HIX}
\end{figure}

Comparing the comprehensibility of the different approaches revealed that the holistic approach had the highest comprehensibility, with a mean HIX value of 15.3 (SD: 3.53). The linguistic-level based approach yielded a mean HIX value of 9.53 (SD: 2.96), whereby the source text had a mean HIX value of 6.04 (SD: 2.84) (see Figure~\ref{fig:HIX}). As mentioned in Section~\ref{sec:design}, the benchmark for a text to be classified as a German Easy Language text is set at 18 points. Therefore, we can conclude that none of the texts that were generated with the linguistic level approach can be classified as German Easy Language texts. In comparison, the holistic approach yielded four texts with a HIX value of at least 18, so  that – according to this criterion – 20\% of the texts could indeed be classified as being easy to understand.

\subsection{Syntactic complexity}
\label{ssec:res:syntacticcomplexity}
\begin{figure*}[!ht]
   \centering
    \includegraphics[width=0.7\textwidth]{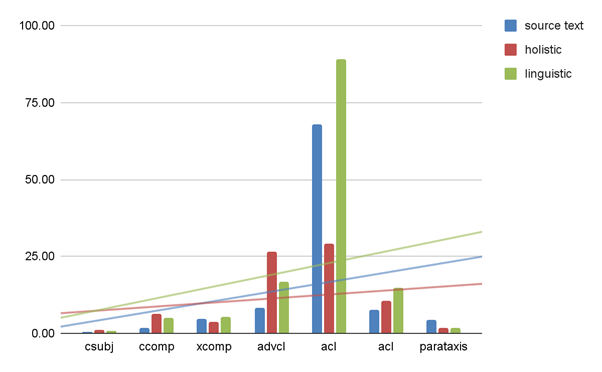}
    \caption{Distribution of syntactically complex dependency relations in the source text and the two simplified variants under analysis.}
     \label{fig:Syntactic}
\end{figure*}

In the next step, we analyse the distribution of the dependency relations across the three subcorpora under analysis. We summarise the results (frequencies normalised per 1000) in Figure~\ref{fig:Syntactic}.

Overall, both simplified text versions seem to have a higher number of complex syntactic relations than the source text. For the latter, we observe higher number for parataxis relations only. Clausal subjects ({\tt csubj}), clausal complements of verbs and adjectives ({\tt ccomp}), as well clauses modifying verbs and adjectives ({\tt advcl}) predominate in the holistic version, whereas subjectless clausal complements ({\tt xcomp}) and clauses modifying nouns ({\tt acl}) prevail in the text version simplified with a linguistic approach. Clauses modifying verbs and adjectives that are in general most frequent amongst all the relations under analysis often include temporal and locative clauses, and clauses that express manner, reason, consequence, alternative or condition. The sentence in the text version simplified with a holistic approach illustrated in example 2a contains two causes of this type: one starting with {\sl wenn} ({\sl if}, condition), and the second with {\sl um} ({\sl so that}, consequence). Both the source text and the text simplified with a linguistic approach are identical (2b) and contain only one {\tt advc} relation expressing condition.

\begin{quote}
2a. \textit{Eltern bekommen auch einen Bonus, wenn sie sich abwechseln, um auf das Baby aufzupassen. [Parents also get a bonus when they take turns taking care of the baby]} (10H)

2b. \textit{Zwei Partnermonate werden zusätzlich als Bonus gewährt, wenn der jeweils andere Elternteil in dieser Zeit seine Erwerbstätigkeit zugunsten der Kindererziehung zeitlich einschränkt oder aussetzt. [Two additional months of parental leave are granted as a bonus when the other parent reduces or suspends their employment during this time for the purpose of child care.]} (10S, 10L)
\end{quote}

An example of the other frequent syntactic relation, i.e. clauses modifying nouns ({\tt acl}), is illustrated in (3). Here we observe a relative clause in the text version simplified with a linguistic approach (3a) and a conditional clause instead in the version simplified with a holistic approach (3b).

\begin{quote}
3a. \textit{Eltern, die vor der Geburt ihres Kindes nicht erwerbstätig waren, erhalten ein Mindestelterngeld von 300 Euro monatlich. [Parents, who were not employed before the birth of their child, receive a minimum parental allowance of 300 euros per month]} (10L)

3b. \textit{Wenn Eltern vor dem Baby nicht arbeiteten, bekommen sie mindestens 300 Euro im Monat. [If parents did not work before the baby, they get at least 300 euros per month.]} (10H).
\end{quote}

Another syntactic construction which is least frequent in the holistic output is {\tt xcomp} (subjectless clauses) in (4).

\begin{quote}
4a. \textit{Falls mehrere Termine gebucht werden, behält sich der Fachdienst das Recht vor, zusätzliche Termine zu löschen, um anderen Bürgern zeitnahe Terminreservierungen zu ermöglichen. [If multiple appointments have been booked, the authority reserves the right to cancel additional appointments to allow other citizens to book appointments in a timely manner.]} (18L).

4b. \textit{Sollten mehrere Termine gebucht werden, behält sich der Fachdienst vor, die weiteren Termine zu löschen, um Terminkapazitäten nicht einzuschränken und anderen Bürgerinnen und Bürgern ebenfalls zeitnahe Terminreservierungen zu ermöglichen. [If multiple appointments have been booked, the authority reserves the right to cancel the additional appointments so as not to restrict appointments capacities and to allow other citizens to book appointments in a timely manner.]}
(18S). 
\end{quote}

In summary, simplified texts turned out to contain less complex syntactic constructions for certain relations only. 

\section{Summary and Discussion}\label{sec:summarydiscussion}
The present paper focused on the feasibility of using ChatGPT for intralingual translation, i.e. translation of administrative texts into German Easy Language. Our results show that in terms of readability, the generated texts are easier than the source texts, however, most of the texts still do not meet the Easy Language standards. In other words, the texts are easier, but not easy enough. Furthermore, the content of the texts was not always correct. However, in terms of correctness, it should be noted that classifying a text as ``incorrect" does not mean that the entire content was incorrect. In most texts that were labelled as incorrect, most of the content was transferred correctly and only one small piece of information was incorrect, or some crucial information was missing, which in turn led to the fact that the message differed from the source text.

All in all, our results allow us to conclude that so far, ChatGPT might be used as a template for professional translators rather than a standalone Easy Language translation tool. The conclusion that in Easy Language translation, human translators are still indispensable is also due to the fact that only parts of the translation can be performed by adhering to simplification rules. Even when all rules are applied, there are still some tasks that require the translator’s specialized knowledge, creativity and understanding and awareness of the target group. 

Therefore, in addition to the text perspective, a functional Easy Language translation also has to focus on the reader and has to be adapted to the reader’s prior knowledge. This for example means that the translator, on the one hand, has to select and prioritize the information for its users and, on the other hand, has to add paraphrases, examples and explanations. As information is processed and retained more easily if presented in a multimodal and multicodal way, the translator also has to include images to reflect, clarify, or exemplify the subject-related information and to highlight core concepts and associations. This shows that even though in Easy Language translation there clearly is a potential for automation, the translation task consists of much more than applying text-based rules. Thus, if translators use ChatGPT to translate texts into German Easy Language, they need to have professional post-editing competences for intralingual translation, such as error detection, research, and correction skills.

However, the more we engaged with the topic, the more we learnt how to get more precise and tailored outputs, i.e., we learnt that other - more appropriate - prompts can improve the comprehensibility of the generated texts. One way to improve the quality of the answers is to assign ChatGPT a role. For example, when telling the tool that it is a translator for Easy Language before asking it to translate a text into Easy Language, it seems that the output is less complex than without the previous role assignment. ChatGPT seems also to deliver more appropriate outcomes if a context is set before asking for a translation. For instance, it may be helpful to ask ChatGPT about German Easy Language rules and then ask for a translation into German Easy Language. A set contextual framework may deliver more appropriate results. Still, the extent to which the versions differ from each other still has to be investigated in a larger-scale study. 

Considering that there are no prompting instructions when opening ChatGPT, we expect that the average user is not aware of these techniques i.e., they do not know that assigning a role or setting a context improves the quality of the output. This highlights the paramount importance of professional competences when using these kind of tools in intralingual Easy Language translation. 

In our future work, we will extend the evaluation techniques applied, as we have focused on those commonly used in German Easy Language research so far. We will also include further automated evaluation and quality estimation methods derived from automatic text simplification. Moreover, we would like to more closely look into different cases of partial correctness mentioned above, where only piece of information was incorrect or missing.

\bibliographystyle{acl_natbib}
\bibliography{ranlp2023}


\end{document}